\documentclass[10pt,twocolumn,letterpaper]{article}

\usepackage[pagenumbers]{cvpr} 

\usepackage{booktabs}
\usepackage{multirow}
\usepackage{color, colortbl}
\usepackage{subcaption}
\usepackage{tabu}
\usepackage{pifont}
\usepackage{makecell}
\usepackage{paralist}
\usepackage{threeparttable}
\usepackage{enumitem}
\usepackage{hhline}
\usepackage{scalerel,xparse}
\usepackage{tcolorbox}

\newcommand{\cmark}{\ding{51}}%
\newcommand{\xmark}{\ding{55}}%

\def\logo{\makebox[22pt][l]{\raisebox{-0.9ex}{\includegraphics[height=30pt]{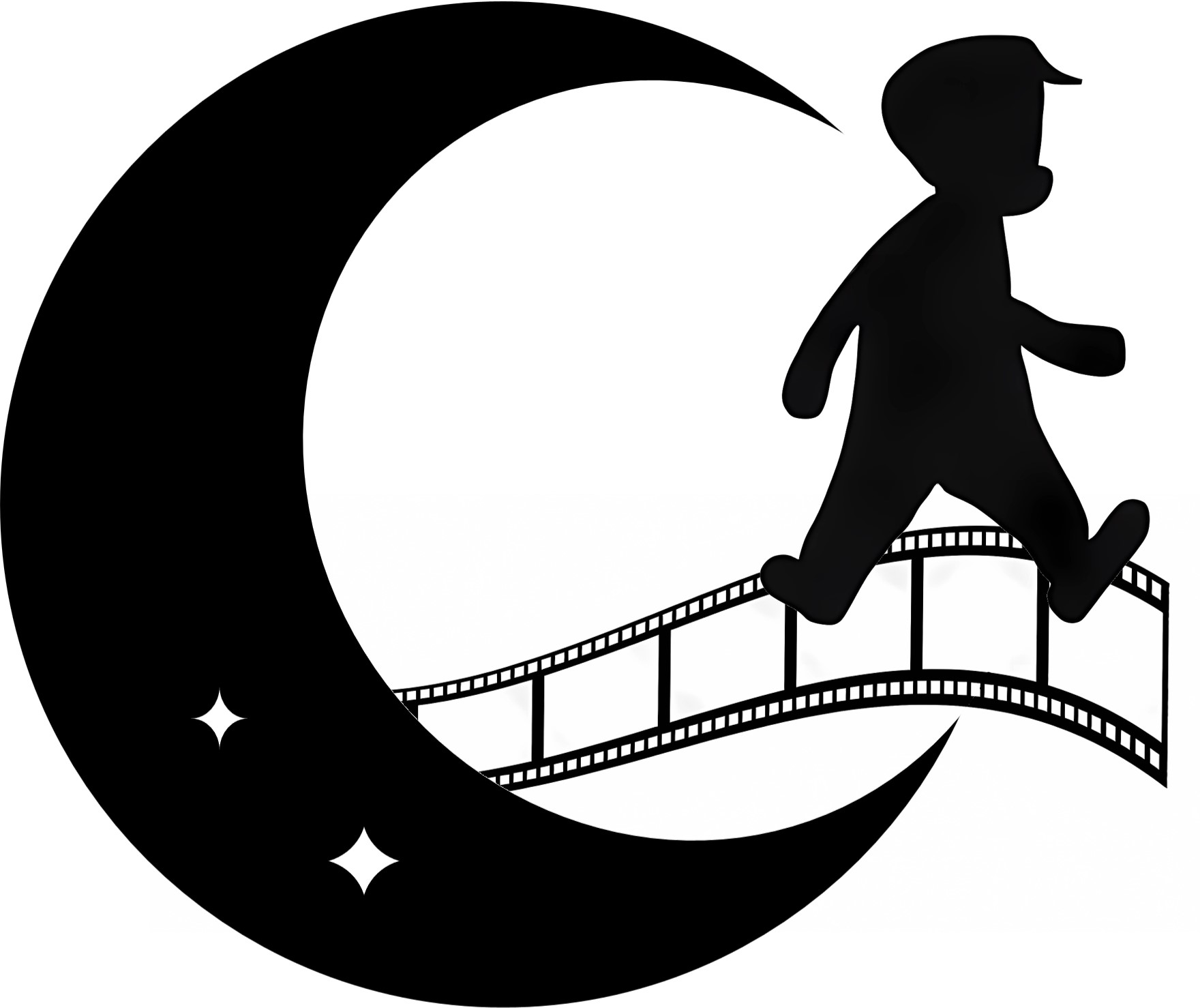}}\hspace{20pt}}}

\newcommand\blfootnote[1]{%
  \begingroup
  \renewcommand\thefootnote{}\footnote{#1}%
  \addtocounter{footnote}{-1}%
  \endgroup
}

\definecolor{cvprblue}{rgb}{0.21,0.49,0.74}
\usepackage[pagebackref,breaklinks,colorlinks,citecolor=cvprblue]{hyperref}

\title{\logo\ \ \ \ \   Vlogger: Make Your Dream A Vlog
}

\author{
    Shaobin Zhuang$^{1,2\spadesuit}$\quad
    Kunchang Li$^{3,4,2\spadesuit}$\quad
    Xinyuan Chen$^{2\heartsuit}$\quad\\
    Yaohui Wang$^{2\heartsuit}$\quad
    Ziwei Liu$^{5}$\quad
    Yu Qiao$^{2,3}$\quad
    Yali Wang$^{1,2,3\heartsuit}$\quad\\
    \footnotesize{$^1$Shanghai Jiao Tong University
    \quad $^2$Shanghai AI Laboratory
    \quad} \\
    \footnotesize{$^3$Shenzhen Institute of Advanced Technology, Chinese Academy of Sciences \quad} \\
    \footnotesize{$^4$University of Chinese Academy of Sciences \quad
    $^5$S-Lab, Nanyang Technological University \quad}
}

\begin{document}
\maketitle
\begin{abstract}

In this work,
we present \textbf{Vlogger},
a generic AI system for generating a \textbf{minute}-level video blog (\textit{i.e.}, vlog) of user descriptions.
Different from short videos with a few seconds,
vlog often contains a complex storyline with diversified scenes,
which is challenging for most existing video generation approaches.
To break through this bottleneck,
our Vlogger smartly leverages Large Language Model (LLM) as Director
and decomposes a long video generation task of vlog into four key stages,
where we invoke various foundation models to play the critical roles of vlog professionals,
including 
(1) Script,
(2) Actor,
(3) ShowMaker,
and 
(4) Voicer.
With such a design of mimicking human beings,
our Vlogger can generate vlogs through explainable cooperation of top-down planning and bottom-up shooting. 
Moreover,
we introduce a novel video diffusion model,
\textbf{ShowMaker},
which serves as a videographer in our Vlogger for generating the video snippet of each shooting scene.
By incorporating Script and Actor attentively as textual and visual prompts,
it can effectively enhance spatial-temporal coherence in the snippet. 
Besides,
we design a concise mixed training paradigm for ShowMaker,
boosting its capacity for both T2V generation and prediction.
Finally,
the extensive experiments show that
our method achieves state-of-the-art performance on zero-shot T2V generation and prediction tasks.
More importantly,
Vlogger can generate over
5-minute
vlogs from open-world descriptions,
without loss of video coherence on script and actor.
The code and model is all available at \href{https://github.com/zhuangshaobin/Vlogger}{https://Vlogger.github.io}.

\end{abstract}   

\begin{figure}[tp]
    \centering
    \includegraphics[width=1.13\linewidth
    ]{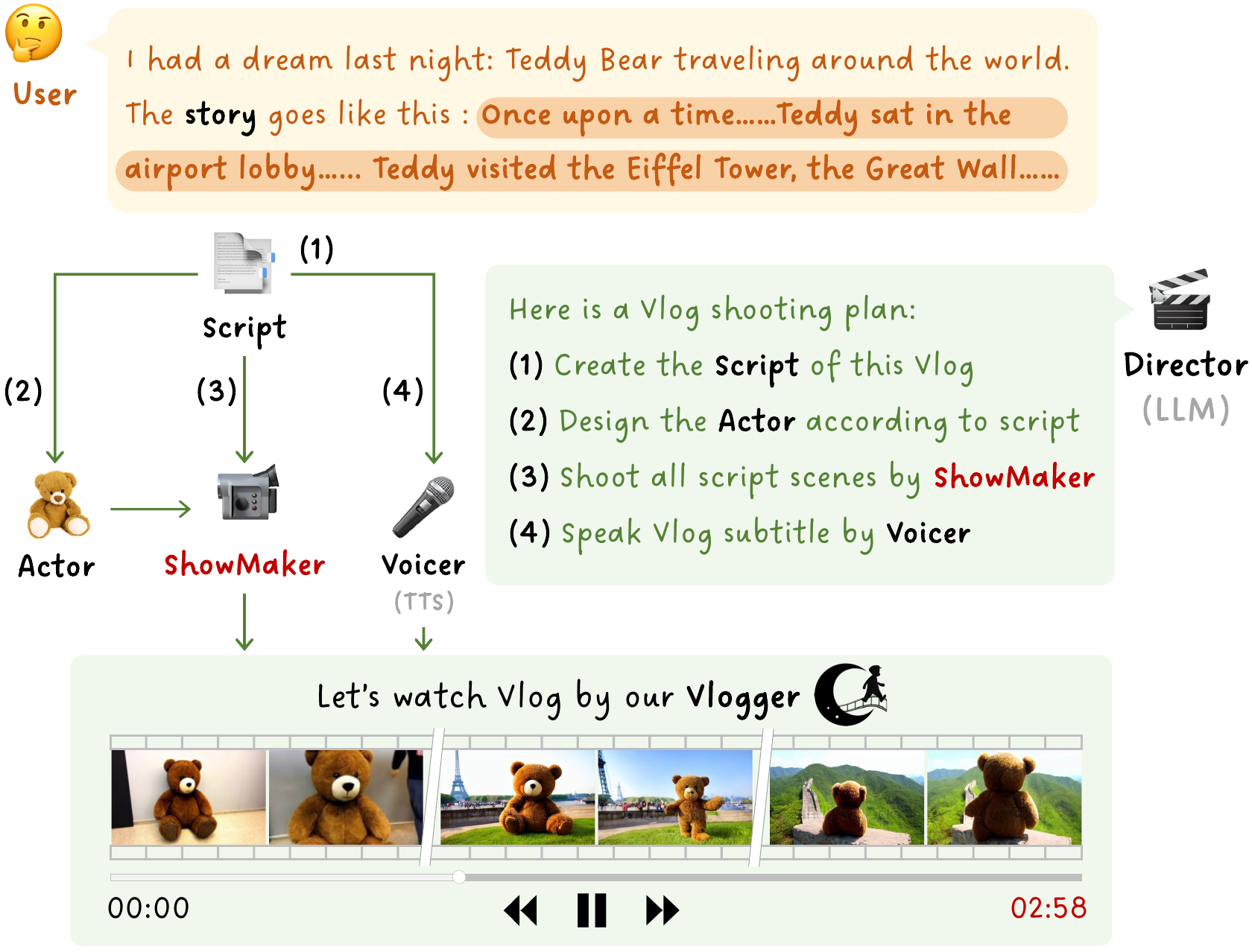}
    \vspace{-0.6cm}
    \caption{
    \textbf{Overview of Vlogger.} 
    Based on the user story, 
    our Vlogger leverages Large Language Model (LLM) as Director,
    and decomposes a minute-long vlog generation task into four key stages with 
    Script, 
    Actor, 
    ShowMaker, 
    and Voicer.
    Moreover,
    ShowMaker is a novel video diffusion model to generate video snippet of each shooting scene,
    with script and actor coherence.
    }
    \label{fig:teaser}
    \vspace{-0.5cm}
\end{figure}

\section{Introduction}
\label{sec:intro}

Vlogs represent a unique form of blogging, distinguished by their utilization of video as the primary medium rather than text.
\blfootnote{$\spadesuit$ Interns at Shanghai AI Laboratory. $\heartsuit$ Corresponding authors.}
Due to more lively expression in the dynamic scenes,
vlog has become one of the most popular online-sharing ways in the digital world. 
In the past two years,
the remarkable success of diffusion models \cite{song2021scorebased, ho2020denoising, ho2021classifierfree} has shown a great impact on video creation \cite{singer2023makeavideo, Ho2022ImagenVH, blattmann2023align, Ge2023PreserveYO, Wang2023ModelScopeTT, Yin2023NUWAXLDO, Zhang2023Show1MP, Wang2023LAVIEHV}.
Hence,
there is a natural question, 
\textit{can we build a generic AI system to generate wonderful vlogs automatically?}

Regrettably, the majority of current video diffusion approaches mainly generate short videos with a few seconds, by temporal adaptation of image diffusion models \cite{singer2023makeavideo, blattmann2023align, Wang2023ModelScopeTT, Zhang2023Show1MP, Wang2023LAVIEHV, Ge2023PreserveYO}.
In contrast,
a vlog typically constitutes a \textit{minute}-level long video in the open world. 
Recently,
there have been some attempts in long video generation \cite{villegas2023phenaki, Yin2023NUWAXLDO}.
However,
these early works either require extensive training on large well-captioned long video datasets \cite{Yin2023NUWAXLDO}
or suffer from noticeable incoherence of shot changes \cite{villegas2023phenaki}. 
Hence,
it is still challenging to generate a minute-level vlog with complex narratives and multiple scenes.

Alternatively,
we notice that
a successful vlog production is a systematical work in our realistic world,
where
the key staffs are involved in
script creation,
actor design, 
video shooting, and 
editing \cite{gao2010vlogging}. 
Drawing inspiration from this,
we believe that,
generating a long-form video blog requires elaborate systematical planning and shooting processes,
rather than simply designing a generative model.

Hence,
we propose a generic AI system for vlog generation in this paper,
namely \textbf{Vlogger},
which can smartly address this difficult task by mimicking vlog professionals with various foundation models in the core steps.
As shown in Fig. \ref{fig:teaser},
we first hire a Large Language Model (LLM) as \textit{Director} (e.g., GPT-4 \cite{OpenAI2023GPT4TR}),
due to its great power of linguistic knowledge understanding.
Given a user story,
this director schedules a four-step plan for vlog generation.
(1) \textit{Script}.
First,
we introduce a progressive script creation paradigm for the LLM Director.
By the coarse-to-fine instructions in this paradigm,
the LLM Director can effectively convert a user story into a script,
which sufficiently describes the story by a number of shooting scenes and their corresponding shooting duration.
(2) \textit{Actor}.
After creating the script,
the LLM Director reads it again to summarize the actors,
and then invokes a character designer (e.g., SD-XL \cite{Podell2023SDXLIL}) to generate the reference images of these actors in the vlog.
(3) \textit{ShowMaker}.
With the guidance of script texts and actor images,
we develop a novel ShowMaker as a videographer,
which can effectively generate a controllable-duration snippet for each shooting scene, with spatial-temporal coherence. 
(4) \textit{Voicer}.
Finally,
the LLM Director invokes a voicer (e.g., Bark \cite{sunoai2023bark}) to dub the vlog with script subtitles.

It should be noted that
our Vlogger overcomes the challenges previously encountered in long video generation tasks.
On one hand, 
it elegantly decomposes the user story into a number of shooting scenes
and designs actor images that can participate in different scenes in the vlog.
In this case,
it can reduce spatial-temporal incoherence of abrupt shot changes,
with explicit guidance of scene texts and actor images.
On the other hand, 
Vlogger crafts individual video snippets for every scene and seamlessly integrates them into a single cohesive vlog. 
Consequently, 
this bypasses the tedious training process with large-scale long video datasets. 
Via such collaboration between top-down planning and bottom-up shooting, 
Vlogger can effectively transform an open-world story into a minute-long vlog.

Furthermore, 
we would like to emphasize that,
\textbf{ShowMaker} is a distinctive video diffusion model designed for generating video snippets of each shooting scene.
From the structural perspective,
we introduce a novel Spatial-Temporal Enhanced Block (STEB) in this model.
This block can adaptively leverage scene descriptions and actor images as textual and visual prompts,
which attentively guide ShowMaker to enhance spatial-temporal coherence of script and actors.
From the training perspective,
we develop a probabilistic mode selection mechanism,
which can boost the capacity of ShowMaker by mixed training of Text-to-Video (T2V) generation and prediction.
More notably,
by sequential combination of generation and prediction mode in the inference stage,
ShowMaker can produce a video snippet with a controllable duration.
This allows our Vlogger to generate a vlog with a preferable duration,
according to the planning of each scene in the script by LLM director.

Finally,
our method achieves the state-of-the-art performance on both zero-shot T2V generation and prediction,
by expensive experiments within the popular video benchmarks.
More importantly,
our Vlogger outperforms the well-known long video generation method,
\textit{i.e.}, 
Phenaki \cite{villegas2023phenaki},
despite utilizing only 66.7\% of the training videos.
Remarkably,
our Vlogger is capable of generating over 5-minute vlogs, 
without loss of script and actor coherence in the video.

\begin{figure*}[t]
    \centering
    \includegraphics[width=1\linewidth]{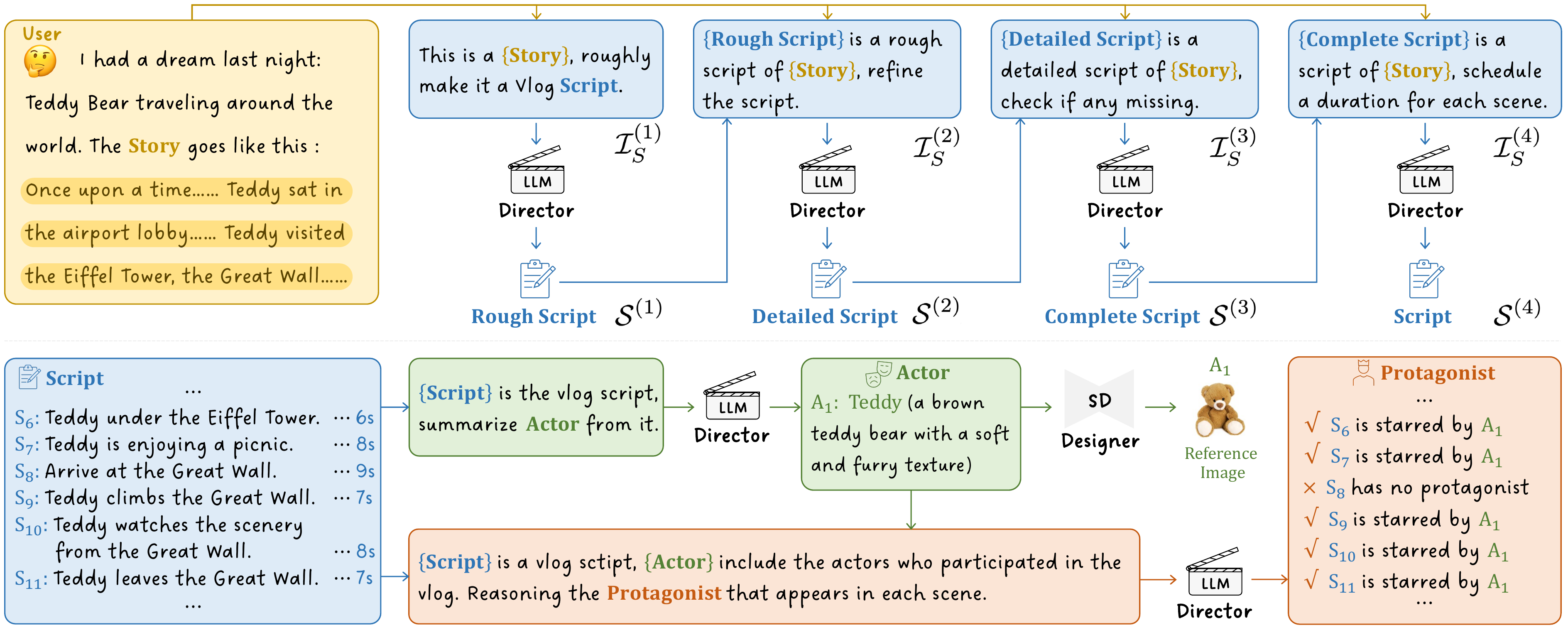}
    \vspace{-0.7cm}
    \caption{
    \textbf{Top-Down Planning.} 
    Through four rounds of dialogue with the LLM, we gradually convert the user story into the final script. 
    Based on this script, we further extract actor reference images,
    and then determine which actor would star in each script scene.
    }
    \label{fig:director}
    \vspace{-0.3cm}
\end{figure*}
\section{Related Works}
\label{sec:related}

\noindent\textbf{Text-to-Video Generation}.
Distinct from conventional unconditional and class-conditional video generation \cite{Wang2020ImaGINatorCS, Wang2019G3ANDA, Wang2021InMoDeGANIM, Chen2020LongTermVP, Clark2019AdversarialVG, Brooks2022GeneratingLV, Yu2022GeneratingVW, Skorokhodov2021StyleGANVAC, Tian2021AGI, Bhagat2020DisentanglingMF, Xie2019MotionBasedGM, Yan2021VideoGPTVG, Ge2022LongVG}, 
T2V generation focuses on automatically converting textual descriptions into videos. 
This is a challenging task,
as it involves understanding text semantics and translating it into video content.
This often requires 
powerful cross-modal algorithms \cite{radford2021learning}, 
large computing resources \cite{Hestness2017DeepLS, Chollet2019OnTM}, 
and extensive video data \cite{schuhmann2022laion5b, bain2021frozen, Wang2023LAVIEHV, Wang2022InternVideoGV, Xue2021AdvancingHV, Schuhmann2021LAION400MOD}.
Based on the success of diffusion models in the Text-to-Image (T2I) generation \cite{Nichol2021GLIDETP, rombach2022highResolution, Ramesh2022HierarchicalTI, saharia2022imagen, Balaji2022eDiffITD, Kusupati2022MatryoshkaRL, Dai2023EmuEI, BetkerImprovingIG},
a series of such works have been recently transferred to T2V generation \cite{singer2023makeavideo, Ho2022ImagenVH, blattmann2023align, Wang2023LAVIEHV, Ge2023PreserveYO, Zhang2023Show1MP, Wang2023VideoFactorySA, Wang2023ModelScopeTT}. 
However,
whether trained from scratch \cite{Ho2022ImagenVH, singer2023makeavideo} or finetuned from T2I model \cite{blattmann2023align, Ge2023PreserveYO, Wang2023LAVIEHV, Wang2023ModelScopeTT, Zhang2023Show1MP, Wang2023VideoFactorySA}, 
most of these approaches mainly work on generating short videos with few seconds from simple descriptions.
In contrast,
our Vlogger can generate a minute-long vlog with complex stories.

\noindent\textbf{Long Video Generation}.
The generation of long videos predominantly relies on parallel \cite{Yin2023NUWAXLDO} or autoregressive \cite{villegas2023phenaki} structures.
However,
these early works still face challenges for vlog generation.
On the one hand,
the parallel manner can relax spatial-temporal content incoherence problems by coarse-to-fine generation.
However, this approach necessitates extensive and laborious training on large, well-annotated long video datasets \cite{Yin2023NUWAXLDO}.
On the other hand,
the autoregressive manner can relax heavy data requirements for training long videos,
by applying short video generation models iteratively with sliding windows.
However,
this solution often suffers from noticeable shot change and long-term incoherence \cite{villegas2023phenaki, chen2023seine, he2023latent, Qiu2023FreeNoiseTL, Wang2023GenLVideoMT}
which becomes problematic when generating vlogs encompassing complex narratives and multiple scenes. 
It is worth mentioning that the community has gradually realized that delegating higher-order reasoning tasks to LLM is of great help to visual tasks \cite{Wu2023VisualCT, gupta2023visual, Li2023VideoChatCV, Zhang2023ControllableTG, Liu2023VisualIT, Zhu2023MiniGPT4EV, Huang2023FreeBloomZT, BetkerImprovingIG}.
Our Vlogger introduces LLM to the field of long video generation,
and effectively addresses the problems of training burden and content incoherence in the previous methods,
by distinct cooperation of top-down planning and bottom-up shooting.

\section{Method}
\label{sec:method}

In this section,
we introduce our Vlogger framework in detail.
First,
we describe how to make a vlog by planning and shooting with our Vlogger.
Then,
we further explain the novel design of ShowMaker.
As the videographer in our Vlogger,
it is critical for video generation in the vlog.

\subsection{Overall Framework of Vlogger}

To generate a minute-level vlog,
our Vlogger leverages LLM as the director,
which can effectively decompose this generation task by four key roles within the planning and shooting stages. 
As shown in Fig. \ref{fig:teaser},
the LLM Director first creates Script and designs Actor in the planning stage.
Based on Script and Actor,
ShowMaker generates a video snippet for each scene in the shooting stage,
and Voicer dubs subtitles of this snippet.
Finally,
one can combine the snippets of all the scenes as a vlog.

\begin{figure*}[t]
    \centering
    \includegraphics[width=0.95\linewidth]{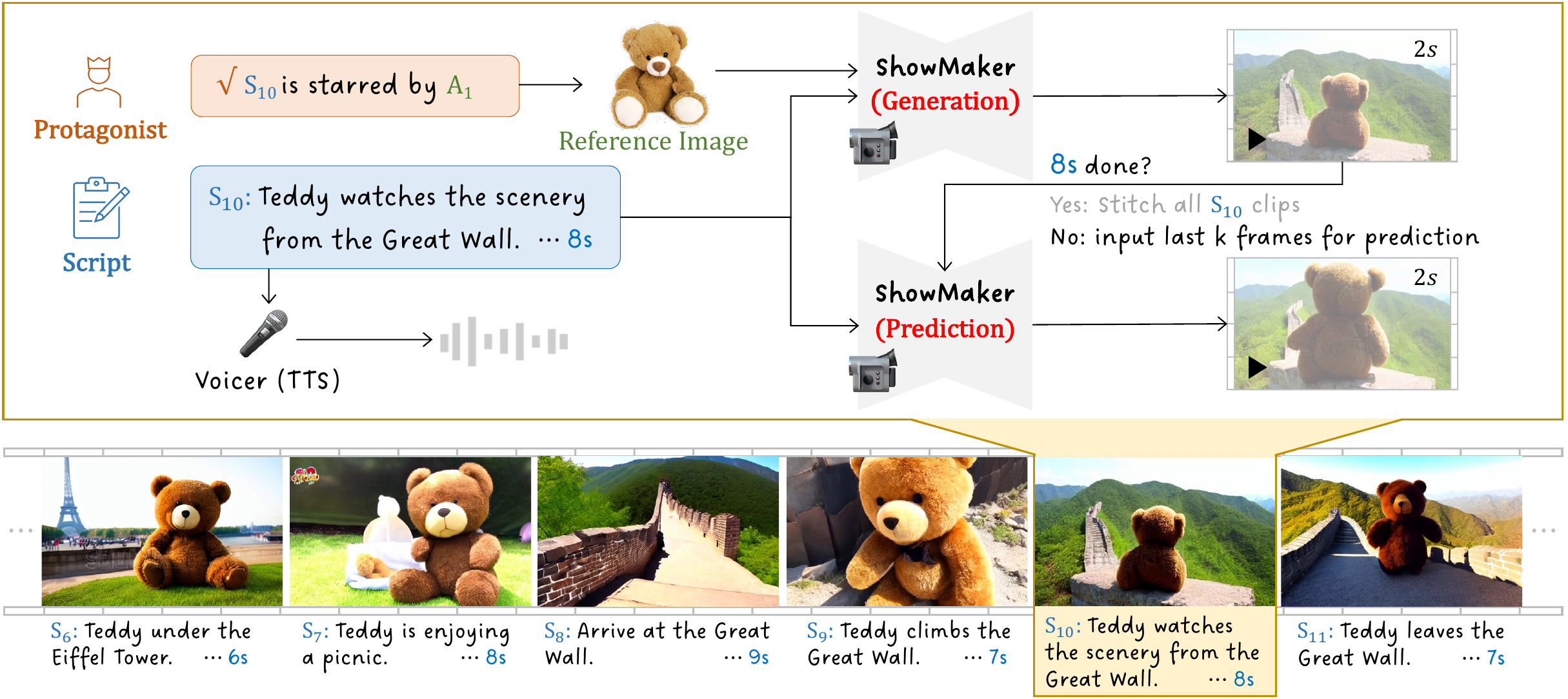}
    \vspace{-0.2cm}
    \caption{
    \textbf{Bottom-Up Shooting.} 
    For each scene,
    ShowMaker can generate the video snippet with script and actor coherence,
    by using script description and actor image as textual and visual prompts.
    Moreover,
    ShowMaker can effectively control the snippet duration,
    by performing generation and prediction sequentially in the inference.  
    Finally,
    we apply a Text-To-Speech (TTS) model as Voicer for dubbing.
    }
    \label{fig:generating}
    \vspace{-0.3cm}
\end{figure*}

\subsubsection{Top-Down Planning}

A vlog often comes from a user story that contains diversified content within many shot changes.
Clearly,
it is challenging to produce a minute-long vlog,
by directly feeding such a complex story into video generation models.
To this end,
we propose to leverage LLM as director and decompose the user story by top-down planning in Fig. \ref{fig:director}.

\textbf{Script Creating}.
First,
we parse the use story into a script,
which describes this story explicitly by a number of shooting scenes.
In this case,
we can generate a video snippet for each shooting scene,
instead of learning a long video tediously from the entire story.
Since LLM has shown an impressive capacity in language understanding \cite{brown2020language, Zeng2022GLM130BAO, Touvron2023LLaMAOA, OpenAI2023GPT4TR, Sun2021ERNIE3L, Bai2023QwenTR, 2023internlm, Chowdhery2022PaLMSL},
we feed the user story into such a director for script generation.
As shown in Fig. \ref{fig:director},
we introduce a progressive creation paradigm,
which can effectively parse the story by coarse-to-fine steps,
\begin{equation}
\mathcal{S}^{(i)}=\text{LLM}(\mathcal{S}^{(i-1)},\mathcal{I}^{(i)}_{S},\mathcal{U}).
\label{eq:script}    
\end{equation}
Given the user story $\mathcal{U}$ and creation instruction $\mathcal{I}^{(i)}_{S}$,
the LLM Director generates the current script $\mathcal{S}^{(i)}$ from the previous one $\mathcal{S}^{(i-1)}$.
More specifically,
there are four steps including
(1) \textit{Rough}.
First,
LLM generates a basic draft of the script from the story.
(2) \textit{Detailed}.
Then,
LLM refines the rough script with story details.
(3) \textit{Completed}.
Next,
LLM checks if the detailed script misses the important parts of the story. 
(4) \textit{Scheduled}.
Finally,
LLM allocates a shooting duration for each scene in the completed script,
according to the scene content.
For convenience,
we denote the final script as $\mathcal{S}$ in the following.
It contains the descriptions of $N$ shooting scenes $\{\mathcal{S}_{1},...,\mathcal{S}_{N}\}$ and their allocated duration $\{\mathcal{T}_{1},...,\mathcal{T}_{N}\}$.
Additionally,
due to the limited pages,
please read the full descriptions of instructions and script in the supplementary doc.

\textbf{Actor Designing}.
After generating the script,
it is time to design actors in the vlog.
As shown in Fig. \ref{fig:director},
we ask the LLM Director to summarize the actor list $\mathcal{A}$ from script $\mathcal{S}$, 
\begin{equation}
\mathcal{A}=\text{LLM}(\mathcal{S},\mathcal{I}_{A}),
\label{eq:actor_des}    
\end{equation}
where 
$\mathcal{I}_{A}$ is the instruction for actor summarization.
Then,
according to actor descriptions,
the LLM Director invokes a designer to generate reference images of these actors $\mathcal{R}$,
\begin{equation}
\mathcal{R}=\text{Stable-Diffusion}(\mathcal{A}).
\label{eq:actor}    
\end{equation}
We choose Stable Diffusion XL \cite{Podell2023SDXLIL} as the character designer,
due to its high-quality generation.
Finally,
based on script $\mathcal{S}$ and actor $\mathcal{A}$,
the LLM Director decides the leading actor (\textit{i.e.}, protagonist) in each shooting scene of the script, 
\begin{equation}
\mathcal{P}=\text{LLM}(\mathcal{S},\mathcal{A},\mathcal{I}_{P}).
\label{eq:actorp}    
\end{equation}
where 
$\mathcal{I}_{P}$ is the instruction for protagonist selection.
The resulting doc $\mathcal{P}$ is aligned with the script $\mathcal{S}$,
where 
$\mathcal{P}_{n}$ describes which actor appears in the scene $\mathcal{S}_{n}$.
For example,
\{$\mathcal{S}_{6}$ \textit{is starred by} $\mathcal{A}_{1}$\} in Fig. \ref{fig:director}.

\begin{figure*}[t]
    \centering
    \includegraphics[width=\textwidth]{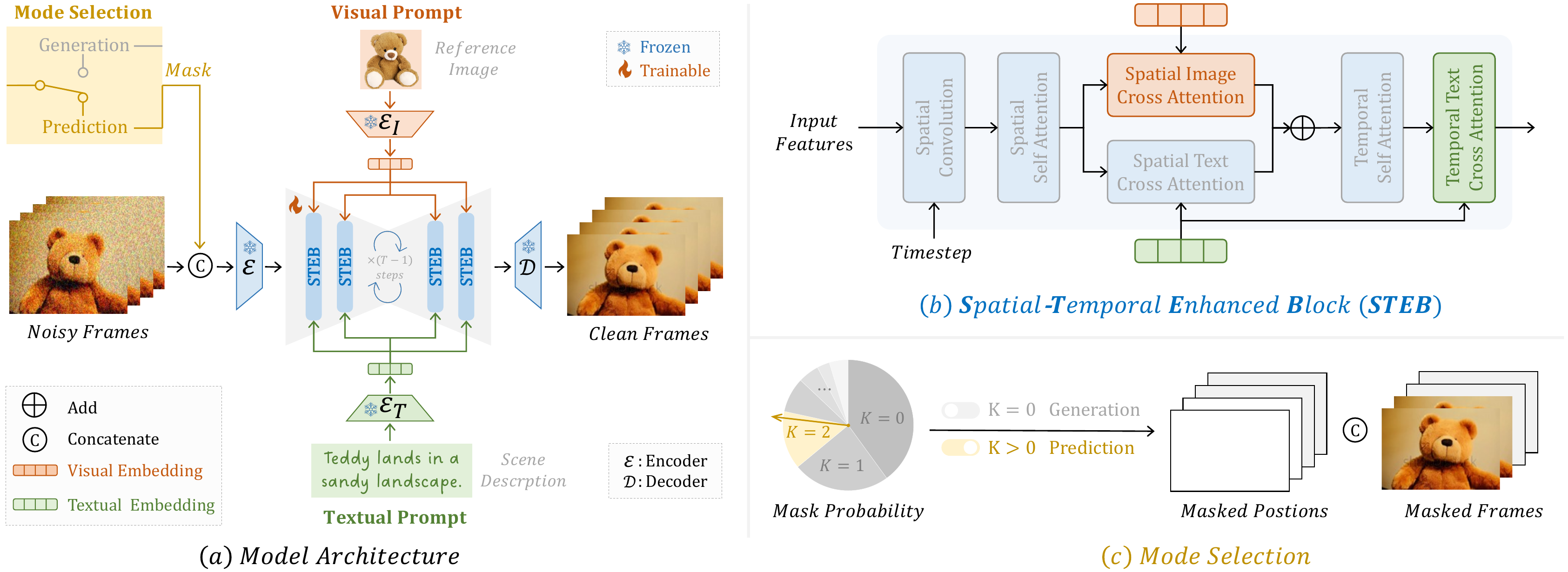}
    \caption{
    \textbf{ShowMaker}. 
    (a) The Overall Architecture.
    (b) Spatial-Temporal Enhanced Block (STEB).
    Via spatial-actor and temporal-text cross attention, 
    our STEB can further enhance actor and script coherence in the snippet.
    (c) Mode Selection. 
    We introduce a mixed training paradigm of T2V generation and prediction,
    via probabilistic selection of masked frames.
    }
    \label{fig:architecture}
    \vspace{-0.3cm}
\end{figure*}

\subsubsection{Bottom-Up Shooting}
\label{sec:Bottom-Up}

Via the top-down planning above,
the LLM Director flexibly decomposes a complex user story into several script scenes 
and designs actor reference image for each scene.
Such a manner largely reduces the difficulty of vlog generation,
since we can generate vlogs by bottom-up shooting,
\textit{i.e.},
we just need to generate the video snippet for each shooting scene and combine all of them as a vlog.

\textbf{ShowMaker Shooting}.
To generate the video snippet of a shooting scene,
we introduce a novel ShowMaker as the videographer,
which is a video diffusion model with two distinct designs.
First,
it is important to maintain spatial-temporal coherence of both script and actor in the generated snippet.
Hence,
our ShowMaker not only takes the scene description $\mathcal{S}_{n}$ as a textual prompt 
but also takes actor image of this scene $\mathcal{R}_{n}$ as a visual prompt.
Second,
each scene is allocated with a shooting duration in the script.
To control the duration of each scene,
our ShowMaker contains two learning modes including generation and prediction in Fig.\ref{fig:generating}.
Specifically,
it starts with the generation mode.
For the shooting scene $n$,
we feed its script description $\mathcal{S}_{n}$ and actor reference image $\mathcal{R}_{n}$ into ShowMaker,
\begin{equation}
\mathcal{C}_{n}^{(1)}=\text{ShowMaker}(\mathcal{N}_{n}^{(1)}~|~\mathcal{S}_{n},\mathcal{R}_{n},\text{Generate}),
\label{eq:showmakergeneration}    
\end{equation}
which generates the first video clip of this scene $\mathcal{C}_{n}^{(1)}$ from the noisy clip $\mathcal{N}_{n}^{(1)}$.
If the duration of this clip is smaller than the allocated duration $\mathcal{T}_{n}$ in the script,
we continue to perform the prediction mode, 
\textit{i.e.},
the last $k$ frames of the current clip $\mathcal{C}_{n}^{(j)}(k)$ are used as context,
when generating the next clip from the noisy input $\mathcal{N}_{n}^{(j+1)}$,
\begin{equation}
\mathcal{C}_{n}^{(j+1)}=\text{ShowMaker}(\mathcal{N}_{n}^{(j+1)}~|~\mathcal{S}_{n},\mathcal{C}_{n}^{(j)}(k),\text{Predict}).
\label{eq:showmakerprediction}    
\end{equation}
Note that,
actor reference images are not necessary in the prediction mode,
since such an actor's appearance has been shown in the current clip $\mathcal{C}_{n}^{(j)}(k)$ that is used as input for prediction.
This prediction procedure stops until the total duration achieves the allocated $\mathcal{T}_{n}$ of this scene.
Subsequently,
we combine all the clips as the video snippet of this scene, \textit{i.e.}, $\mathcal{C}_{n}=\{\mathcal{C}_{n}^{(1)},...,\mathcal{C}_{n}^{(J)}\}$.

\textbf{Voicer Speaking}.
To enhance the completeness of the vlog,
we apply a Text-To-Speech model (e.g., Bark \cite{sunoai2023bark}) as a Voicer,
which converts the scene description $\mathcal{S}_{n}$ into the corresponding audio $\mathcal{O}_{n}=\text{Bark}(\mathcal{S}_{n})$.
Finally,
we add this audio to the corresponding video snippet
$\mathcal{V}_{n}=\mathcal{O}_{n}\oplus\mathcal{C}_{n}$,
and
combine all the sounded video snippets as a complete vlog,
\textit{i.e.},
$\mathcal{V}=\{\mathcal{V}_{1},...,\mathcal{V}_{N}\}$.

\subsection{ShowMaker}

As discussed in Section \ref{sec:Bottom-Up},
ShowMaker plays a critical role in generating video snippet of a shooting scene.
In this work,
we introduce a text-to-video diffusion model for it.
It follows the style of latent diffusion model \cite{rombach2022highResolution}.
In the diffusing stage,
we add Gaussian noise progressively to the latent code of a training snippet.
In the denoising stage,
we reconstruct the latent code from the noisy latent code at any iteration step.
For simplicity,
we just show the denoising stage in Fig.\ref{fig:architecture} (a).
First,
we forward a noisy training snippet into the encoder of the autoencoder to extract its latent code.
Then,
we feed this into a denoising U-Net \cite{Ronneberger2015UNetCN} to learn the clean latent code.
Finally,
we leverage the decoder to reconstruct the original snippet with the clean latent code.

But, compared with the existing video diffusion models \cite{Ho2022VideoDM, blattmann2023align, Wang2023LAVIEHV, Ho2022ImagenVH, Ge2023PreserveYO},
our novel ShowMaker contains two distinct designs,
in terms of 
model structure (\textit{i.e.}, Spatial-Temporal Enhanced Block) in  Fig.\ref{fig:architecture} (b),
and
training paradigm (\textit{i.e.}, Mode Selection) in Fig.\ref{fig:architecture} (c).

\subsubsection{Spatial-Temporal Enhanced Block (STEB)}
\label{sec:STEB}

To reconstruct the clean latent code of a training video snippet,
each block in the denoising U-Net consists of both spatial and temporal operations in previous works \cite{Wang2023LAVIEHV}.
First,
spatial operations encode the feature of each frame separately in the snippet.
Typically,
three operations are inherited from text-to-image generation approaches \cite{saharia2022imagen, Ramesh2022HierarchicalTI, rombach2022highResolution, Balaji2022eDiffITD, Nichol2021GLIDETP},
including 
spatial ConVolution (CV), 
spatial Self Attention (SA),
and 
spatial Cross Attention (CA),
\begin{align}
\mathcal{X}_{cv}&=\text{CV-Spatial}(\mathcal{X}_{in}, \mathcal{F}_{t}),
\label{eq:steb1}\\    
\mathcal{X}_{sa}&=\text{SA-Spatial}(\mathcal{X}_{cv}),
\label{eq:steb2}\\
\mathcal{X}_{ca}&=\text{CA-Spatial-Text}(\mathcal{X}_{sa},\mathcal{S}_{n}),
\label{eq:steb3} 
\end{align}
where
$\mathcal{X}_{in}$ is the noisy feature of the training snippet,
and $\mathcal{F}_{t}$ is the positional embedding of the iteration step $t$.
To guide denoising with the given text, 
we use the scene description $\mathcal{S}_{n}$ as Key and Value of cross attention in Eq. (\ref{eq:steb3}).

However,
we notice that 
such spatial encoding does not consider actors.
Hence,
it inevitably suffers from actor incoherence when generating a snippet.
To tackle this problem,
we introduce a spatial image cross attention,
\begin{equation}
\mathcal{Y}_{ca}=\text{CA-Spatial-Actor}(\mathcal{X}_{sa},\mathcal{R}_{n}).
\label{eq:steb4} 
\end{equation}
For a shooting scene $\mathcal{S}_{n}$,
we leverage the protagonist doc (Eq. \ref{eq:actorp}) to find the corresponding actor in this scene.
Then,
we leverage actor reference image $\mathcal{R}_{n}$ as the visual context of spatial cross attention.
Subsequently,
we enhance spatial embedding with the complementary guidance of both script and actor,
\textit{i.e.},
$\mathcal{Z}_{se}=\mathcal{X}_{ca}+\beta\mathcal{Y}_{ca}$ with a weight $\beta$.

After spatial encoding,
it is time to learn the correlation across frames in the snippet.
Hence,
the typical operation is to perform self attention along the temporal dimension,
\begin{equation}
\mathcal{Z}_{sa}=\text{SA-Temporal}(\mathcal{Z}_{se}).
\label{eq:steb5} 
\end{equation}
However,
such a temporal operation does not take the constraints of scene text into account.
It often leads to text incoherence when generating video snippets.
Hence,
we introduce a temporal text cross attention,
\begin{equation}
\mathcal{Z}_{ca}=\text{CA-Temporal-Text}(\mathcal{Z}_{sa},\mathcal{S}_{n}),
\label{eq:steb6} 
\end{equation}
where
we leverage the scene description $\mathcal{S}_{n}$ as the textual context of temporal encoding.
Via spatial-actor (Eq. \ref{eq:steb4}) and temporal-text (Eq. \ref{eq:steb6}) cross attention,
our STEB can further enhance actor and script coherence in the snippet.

\subsubsection{Mixed Training Paradigm with Mode Selection}

As discussed in Section \ref{sec:Bottom-Up},
ShowMaker aims to generate a video snippet with the allocated duration in the script.
To this goal,
it leverages the combination of both generation and prediction modes in the inference stage.
Next,
we explain how to train our ShowMaker to learn these modes.

\begin{table}[tp]
\centering
\begin{minipage}[t]{0.49\linewidth}
    \vspace{0pt}
    \centering
    \setlength\tabcolsep{7.0pt}
    \resizebox{0.986\textwidth}{!}{
        \begin{tabular}{l c}
        \toprule
        \textbf{Method} & \textbf{FVD ($\downarrow$)}  \\
        \midrule        
        VideoFactory~\cite{Wang2023VideoFactorySA} & 410.00 \\
        Make-A-Video~\cite{singer2023makeavideo} & 367.23 \\
        PYoCo~\cite{Ge2023PreserveYO} & 355.19 \\
        \midrule
        \textbf{Ours} & \textbf{292.43} \\
        \bottomrule
    \end{tabular}
    }
    \vspace{0.2cm}
    \subcaption{Hand-crafted prompt.}
    \label{tab:ucf_manual}
\end{minipage}
\begin{minipage}[t]{0.49\linewidth}
    \vspace{0pt}
    \centering
    \centering
    \setlength\tabcolsep{3.8pt}
    \resizebox{1.04\textwidth}{!}{
        \begin{tabular}{l c}
        \toprule
        \textbf{Method} & \textbf{FVD ($\downarrow$)}  \\
        \midrule
        CogVideo (Chinese)~\cite{hong2023cogvideo} & 751.34 \\
        CogVideo (English)~\cite{hong2023cogvideo} & 701.59 \\
        MagicVideo~\cite{Zhou2022MagicVideoEV} & 699.00 \\
        Video LDM~\cite{blattmann2023align} & 550.61 \\
        \midrule
        \textbf{Ours} & \textbf{525.01} \\
        \bottomrule
    \end{tabular}
    }
    \vspace{-0.1cm}
    \subcaption{Class label.}
    \label{tab: ucf_class}
\end{minipage}
\vspace{-0.3cm}
\caption{\textbf{Zero-shot comperison with the state-of-the-art methods on UCF-101}.
Hand-crafted prompt from \cite{Ge2023PreserveYO}.
}
\vspace{-0.3cm}
\label{tab:ucf}
\end{table}

As shown in Fig.\ref{fig:architecture} (c),
we design a mode selection mechanism,
which selects $k$ frames of the clean snippet as the context of the noisy snippet.
The $k$$=$$0$ setting refers to the generation mode 
since there is no context from the clean snippet.
Alternatively,
the $k$$>$$0$ setting refers to the prediction mode,
since $k$ frames of the clean snippet are already available.
The goal is to generate the rest frames of the clean snippet from the noisy one.
To integrate both modes into training,
we design a probabilistic manner to select $k$,
\begin{align}
\begin{split}
\text{P}(k)=\left\{
\begin{aligned}
\alpha^{k}-\alpha^{k+1},~~~~~~& k\in\lbrack0,m) \\
\alpha^{k},~~~~~~& k=m
\end{aligned}
\right.
\end{split}
\label{eq:alpham} 
\end{align}
where 
$\text{P}(k)$ is the selection probability distribution of $k$.
Moreover,
$0$$<$$\alpha$$<$$1$ and $0$$\leq$$m$ are the manual parameters,
which respectively control the mode selection tendencies of $\text{P}(k)$ and the maximum number of preserved frames. 

After determining $k$,
we introduce a frame mask $\mathcal{M}_{k}$ on the latent code of the clean snippet $\mathcal{X}_{clean}$,
\begin{equation}
\mathcal{X}_{k}=\mathcal{X}_{clean}\odot\mathcal{M}_{k}, \label{eq:ms1}
\end{equation}
where
we only preserve $k$ frames and mask the rest of the frames.
Then,
we concatenate the mask $\mathcal{M}_{k}$ with the latent code of snippet $\mathcal{X}_{noise}$ and $\mathcal{X}_{k}$,
\begin{equation}
\mathcal{X}_{in}=\text{Concat}(\mathcal{X}_{noise},~\mathcal{X}_{k},~\mathcal{M}_{k}).\label{eq:ms3}  
\end{equation}
This produces the input feature $\mathcal{X}_{in}$ for training the denoising U-Net in Section \ref{sec:STEB}.
Via such a concise probabilistic manner,
we can effectively integrate both generation ($k$$=$$0$) and prediction ($k$$>$$0$) modes in the training procedure.

\begin{table}[t]
    \centering
    \setlength\tabcolsep{9pt}
    \resizebox{1\linewidth}{!}{%
    \begin{tabular}{l  c c c}
        \toprule
        \textbf{Method} &  \textbf{Zero-Shot} & \textbf{Pre-training Videos} & \textbf{FID ($\downarrow$)}  \\
        \midrule        
        T2V~\cite{li2017video} & \xmark & \xmark & 82.13 \\
        SC~\cite{balaji2019Conditional} & \xmark & \xmark & 33.51 \\
        TFGAN~\cite{balaji2019Conditional} & \xmark & \xmark & 31.76 \\
        NUWA~\cite{wu2022nwa} & \xmark & 0.97M & 28.46 \\
        \midrule
        Phenaki~\cite{villegas2023phenaki} & \cmark & 15M & 37.74 \\
        \textbf{Ours} & \cmark & 10M & \textbf{37.23} \\
        \bottomrule
        \vspace{-2em}
    \end{tabular}
    }
    \caption{\textbf{Comperison with the state-of-the-art methods on Kinetics-400}. 
Both under a zero-shot setting, our FID is better than that of Phenaki, 
with only 66.7\% pre-training videos.}
    \vspace{-1em}
    \label{tab:k400}
\end{table}
\begin{table}[t]
    \centering 
    \setlength\tabcolsep{3pt}
    \resizebox{1.0\linewidth}{!}{%
    \begin{tabular}{l  c c c}
        \toprule
        \textbf{Method} &  \textbf{Zero-Shot} & \textbf{CLIPSIM ($\uparrow$)} & \textbf{CLIP-FID ($\downarrow$)}  \\
        \midrule
        GODIVA~\cite{Wu2021GODIVAGO} & \xmark & 0.2402 & - \\
        N\"{U}WA~\cite{wu2022nwa} & \xmark & 0.2439 & 47.68 \\
        CogVideo (Chinese)~\cite{hong2023cogvideo} & \cmark & 0.2614 & 24.78 \\
        CogVideo (English)~\cite{hong2023cogvideo} & \cmark & 0.2631 & 23.59 \\
        Video LDM~\cite{blattmann2023align} & \cmark & 0.2929 & - \\
        Make-A-Video~\cite{singer2023makeavideo} & \cmark & 0.3049 & 13.17 \\
        \color{gray}{PYoCo}~\cite{Ge2023PreserveYO} & \color{gray}{\cmark} & \color{gray}{-} & \color{gray}{10.21} \\
        \midrule
        \textbf{Ours} & \cmark & 0.2908 & \textbf{12.67} \\
        \bottomrule
        \vspace{-2em}
    \end{tabular}
    }
    \caption{\textbf{Comperison with the state-of-the-art methods on MSR-VTT}. PYoCo \cite{Ge2023PreserveYO} sample 59,794 videos for CLIP-FID (noted in {\color{gray}{gray}}) while others only sample less than 3,000 videos.}
    \vspace{-1.2em}
    \label{tab:msrvtt}
\end{table}
\section{Experiments}
\label{sec:exper}

\textbf{Datasets}.
To make state-of-the-art comparison,
We conduct zero-shot evaluation on the popular video benchmarks,
\textit{i.e.},
UCF-101 \cite{Soomro2012UCF101AD}, 
Kinetics-400 \cite{Kay2017TheKH}, 
and 
MSR-VTT \cite{Chen2021TheMT}.
UCF-101 contains videos of 101 action categories.
Following \cite{unterthiner2019accurate},
we use FVD to evaluate the distance between the generated video and the real video.
Kinetics-400 is a dataset comprising videos of 400 action categories. 
Following \cite{li2017video, balaji2019Conditional, wu2022nwa, villegas2023phenaki},
we use FID \cite{heusel2017gans} to assess the performance of video generation.
MSR-VTT is a video dataset with open-vocabulary captions,  
wherein CLIPSIM \cite{Wu2021GODIVAGO} and CLIP-FID \cite{parmar2022aliased} are commonly employed for evaluating the T2V generation.
Moreover,
since these existing benchmarks either have a small number of testing videos or contain only action labels without complex descriptions,
we propose to collect an evaluation benchmark for ablation studies.
It is called as Vimeo11k,
where we collect 11,293 open-world videos along with their captions from 10 mainstream categories in Vimeo.
To our best knowledge,
it is the largest testing benchmark for zero-shot video generation. 
More details and experiments can be found in the supplementary doc.

\noindent\textbf{Implementation Details}. 
\textbf{(1) Director \& Script \& Actor \& Voicer. }
We choose GPT-4 \cite{OpenAI2023GPT4TR} as our LLM director to generate script.
The specific instructions can be found in the supplementary doc.
Then we employ Stable Diffusion XL \cite{Podell2023SDXLIL} and Bark \cite{sunoai2023bark} as our designer and voicer to generate reference actor images and convert scripts into speech.
\textbf{(2) ShowMaker. }
We choose SD-1.4 \cite{rombach2022highResolution} as our base model, 
and follow \cite{singer2023makeavideo} to add temporal self attention.
Then,
we add our temporal text cross attention on top of each temporal self attention.
We expand the input channel of the conv-in layer of U-Net from 4 to 9, 
so that model can take the concatenated feature in Eq. (\ref{eq:ms3}) as input.
We use zero initialization for newly added channels.
We use CLIP ViT-L/14 \cite{radford2021learning} as text encoder $\varepsilon_{T}$,
VQVAE \cite{Esser2020TamingTF} as autoencoder consisting of $\varepsilon$ and $\mathcal{D}$.
Besides,
we use OpenCLIP ViT-H/14 \cite{ilharco2021openclip} as image encoder $\varepsilon_{I}$,
and add spatial image cross attention in the STEB block.
The diffusion step $T$ is set to 1000 as \cite{rombach2022highResolution}.
For training, 
we set the parameter of mode selection as $\alpha=0.6$ and $m=6$ in Eq. (\ref{eq:alpham}),
and $\beta$ is set to 1.
Following \cite{Ho2022ImagenVH, Wang2023LAVIEHV, Ho2022VideoDM, Dandi2019JointlyTI, villegas2023phenaki},
we employ publicly available image dataset (\textit{i.e.}, Laion400M \cite{Schuhmann2021LAION400MOD}) and video dataset (\textit{i.e.}, WebVid10M \cite{bain2021frozen}) for joint training.
More training details can be found in the supplementary doc.

\begin{figure}[t]
    \centering
    \includegraphics[width=0.9\linewidth
    ]{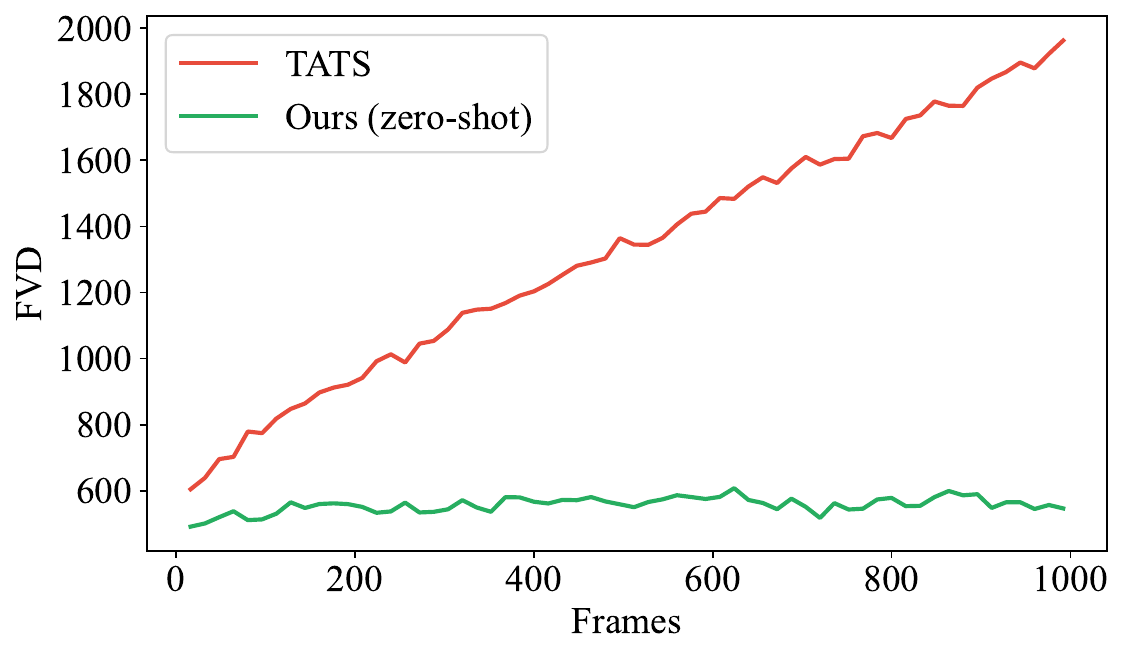}
   \vspace{-0.25cm}
    \caption{
    \textbf{Long video generation.}
    Comparison on UCF-101 (1000 frames). 
    The lower FVD, the better generation performance.
    }
    \label{fig:ucf_1024}
    \vspace{-0.3cm}
\end{figure}

\begin{table}[t]
    \centering
    \setlength\tabcolsep{4.5pt}
    \resizebox{1\linewidth}{!}{%
    \begin{tabular}{l c c c}
        \toprule
        \textbf{Method} & \textbf{CLIP-I ($\uparrow$)} & \textbf{CLIP-T ($\uparrow$)} & \textbf{Vlog Duration ($\uparrow$)}  \\
        \midrule
        w/o \small{autoregressive}~\cite{Zhu2023MovieFactoryAM} & 0.5683 & 0.2752 & 1min03s \\
        only \small{autoregressive}~\cite{villegas2023phenaki} & 0.5642 & 0.2535 & 3min58s \\
        \textbf{Ours} & \textbf{0.6294} & \textbf{0.2756} & \textbf{3min58s} \\
        \bottomrule
        \vspace{-1.75em}
    \end{tabular}
    }
    \caption{\textbf{Ablation for generation process}.
    \cite{Zhu2023MovieFactoryAM} cannot generate a long video in a single scene and both \cite{Zhu2023MovieFactoryAM, villegas2023phenaki} can't refer to images.
    }
    \vspace{-1em}
    \label{tab:framework_ablation}
\end{table}

\subsection{Comparison with state-of-the-art}

Tab. \ref{tab:ucf} shows that,
no matter whether the input text is the class label or hand-crafted prompt, 
our method achieves the best FVD performance of zero-shot video generation on UCF-101.
Tab. \ref{tab:k400} shows that,
compared to Phenaki \cite{villegas2023phenaki},
our method achieves a better FID performance of zero-shot setting on Kinetics-400, 
but only using 66.7\% training videos.
Furthermore, our generated videos have a resolution of $320$$\times$$512$, which is higher than Phenaki's $256$$\times$$256$.
Tab. \ref{tab:msrvtt} shows that,
our method achieves remarkably competitive FID and CLIPSIM performance on MSR-VTT.
Furthermore, 
Fig. \ref{fig:ucf_1024} illustrates that,
we significantly surpass TATS \cite{ge2022long} (\textit{i.e.}, the only open-source long video generation model within our knowledge),
for generating 1000-frame videos on UCF-101.
Moreover,
our method does not encounter the issue of TATS,
where the video quality continuously declines as the number of frames increases.
It is noteworthy that,
our method achieves this performance by zero-shot generation,
without any finetuning on UCF-101.

\subsection{Ablation Study}

\textbf{Vlog Generation Process}.
Tab. \ref{tab:framework_ablation} presents a comparison between different generation processes. 
For a fair comparison, 
we use the same scripts and employ our ShowMaker as videographer. 
We compare three approaches including
no autoregression at all as MovieFactory \cite{Zhu2023MovieFactoryAM}, 
fully autoregressive processes in one go as Phenaki \cite{villegas2023phenaki}, 
and 
the generation process of our Vlogger.
Specifically,
We use GPT-4 to generate five stories and go through the Vlogger’s planning process to get five vlog scripts.
We set 20 different random seeds,
so that each of the above generation processes generates 20 different vlogs for each script.
We use CLIP-I, CLIP-T \cite{Ye2023IPAdapterTC} and video duration to evaluate the quality of the generated vlogs. 
The results demonstrate that,
even using the same script and videographer, 
our Vlogger outshines other existing frameworks for preferable vlog generation. 
More details can be found in the supplementary doc.

\begin{figure}[t]
    \centering
    \includegraphics[width=0.9\linewidth
    ]{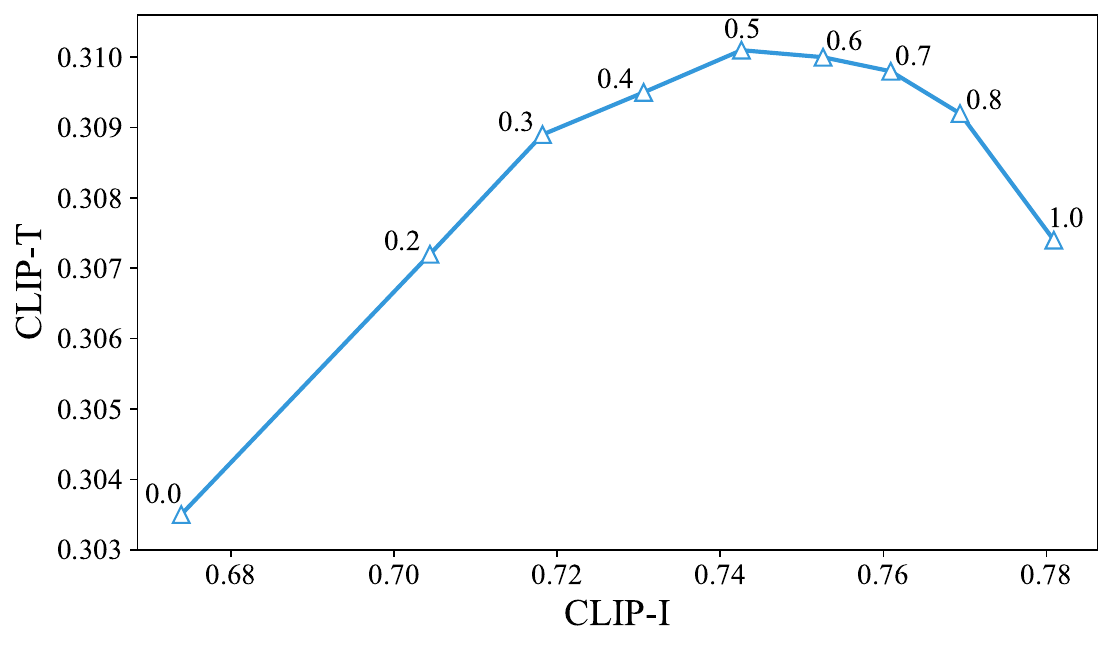}
   \vspace{-0.25cm}
    \caption{
    \textbf{Ablation for spatial image cross attention.} 
    The values around a point on the curve correspond to the $\beta$.
    While $\beta$$=$$0$ means the original network without spatial image cross attention.
    }
    \label{fig:ica}
    \vspace{-0.3cm}
\end{figure}

\begin{figure*}[t]
    \centering
    \includegraphics[width=1\textwidth]{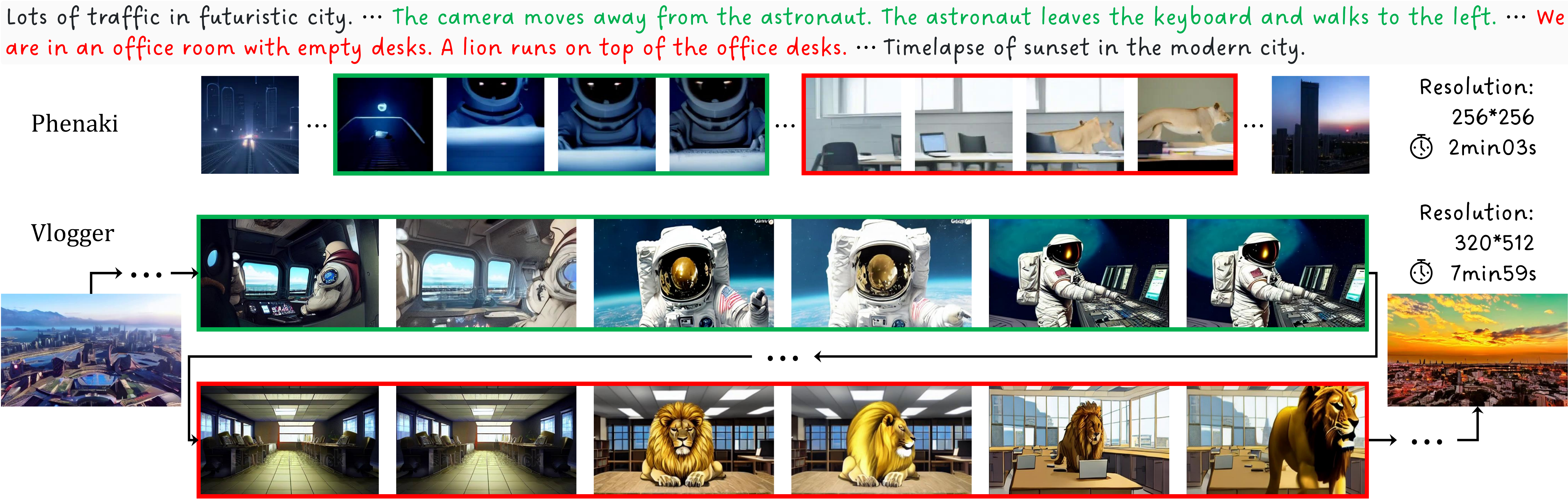}
    \caption{
    \textbf{Qualitative comparison with the state-of-the-art methods on long video generation.}
    The story and long video of Phenaki \cite{villegas2023phenaki} are available at
    \href{https://phenaki.github.io/}{phenaki.github.io}. 
    The scene diversity and picture quality generated by Vlogger are significantly better than Phenaki.
    }
    \label{fig:movie_case}
\end{figure*}

\begin{figure*}[t]
    \centering
    \includegraphics[width=1\textwidth]{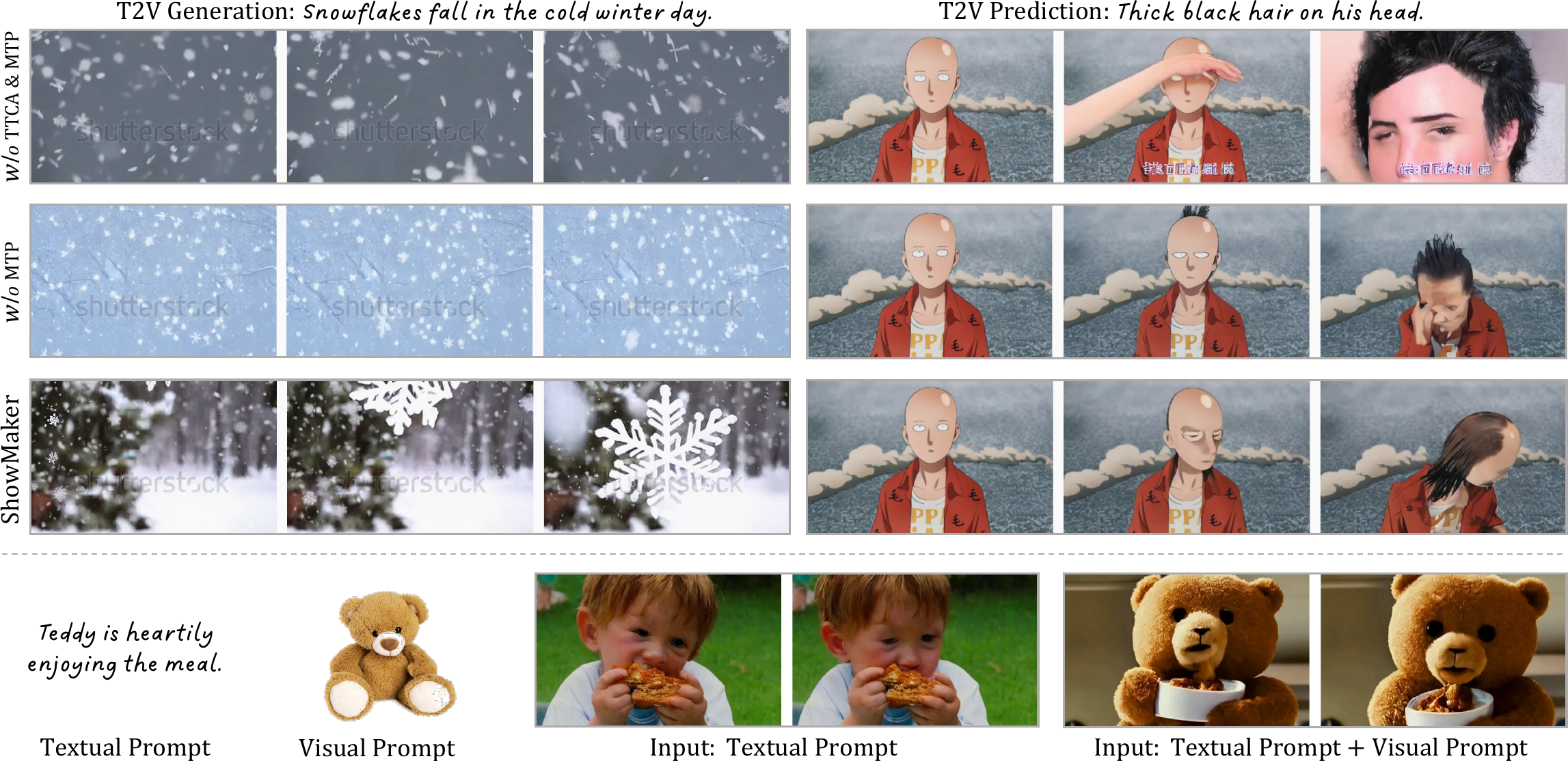}
    \caption{
    \textbf{Qualitative ablation for STEB and training paradigm.}
     ``TTCA'' and ``MTP'' refer to temporal the text cross attention and mixed training paradigm respectively. The sample without a visual prompt corresponds to the model without spatial image cross attention.
     }
    \label{fig:short_case}
    \vspace{-0.2cm}
\end{figure*}

\begin{table}[t]
    \centering
    \setlength\tabcolsep{6pt}
    \resizebox{0.86\linewidth}{!}{%
    \begin{tabular}{c c c c c c}
        \toprule
        \textbf{TTCA} & \textbf{MTP} & \textbf{0/16 ($\downarrow$)} & \textbf{1/15 ($\downarrow$)} & \textbf{3/13 ($\downarrow$)} & \textbf{5/11 ($\downarrow$)}  \\
        \midrule
        \xmark & \xmark & 348.36 & 217.39 & 297.37 & 246.56 \\
        \cmark & \xmark & 333.63 & 178.27 & 230.56 & 193.52 \\
        \cmark & \cmark & \textbf{257.70} & \textbf{123.51} & \textbf{118.02} & \textbf{109.31} \\
        \bottomrule
        \vspace{-1.75em}
    \end{tabular}
    }
    \caption{\textbf{Ablation for temporal text cross attention and mixed training paradigm}. ``TTCA'' and ``MTP'' refer to temporal text cross attention and mixed training paradigm respectively. ``3/13'' denotes predicting the following 13 frames after being given 3 frames of a ground truth video, and we evaluate the FVD between the newly generated 13 frames and the corresponding frames in the ground truth. 
    ``0/16'', ``1/15'', ``5/11'' follow this pattern.
    }
    \vspace{-0.5cm}
    \label{tab:vimeo}
\end{table}

\noindent\textbf{Spatial-Temporal Enhanced Block}.
Fig. \ref{fig:ica} and Tab. \ref{tab:vimeo} evaluate two important operations in our STEB block,
\textit{i.e.}, 
spatial image cross attention and temporal text cross attention.
With spatial image cross attention in Fig. \ref{fig:ica}, 
the CLIP-I is significantly improved as $\beta$ increased, 
while the CLIP-T reached its maximum at $\beta$$=$$0.5$ on COCO2017 \cite{Lin2014MicrosoftCC}.
With temporal text cross attention (TTCA) in Tab. \ref{tab:vimeo}, 
both T2V generation (0/16) and prediction (1/15, 3/13 and 5/11) show a significant improvement on Vimeo11k.

\noindent\textbf{Mixed Training Paradigm}.
Tab. \ref{tab:vimeo} also presents a model comparison by incorporating our mixed training paradigm or the random mask training method in \cite{chen2023seine, voleti2022mcvd, Hoppe2022DiffusionMF, Fu2022TellMW, Yu2022MAGVITMG, Chang2022MaskGITMG}. 
The results show that the mixed training paradigm is effective in enhancing the model's T2V generation and prediction capabilities.
More details in the supplementary doc.

\subsection{Visualization}

First,
we visualize long video generation,
by comparison with the well-known Phenaki \cite{villegas2023phenaki}.
Note that,
since Phenaki does not have the open-sourced codes,
we alternatively use the demo shown in its official website.
Specifically,
we feed the same story description into our Vlogger.
As shown in Fig. \ref{fig:movie_case}, 
our Vlogger shows superior and more diverse video content,
compared to Phenaki.
Moreover,
our Vlogger sufficiently exhibits the story with a much longer duration (\textit{i.e.}, 7min59s),
according to our LLM-created script.

We further visualize the ablation of T2V video generation and prediction by ShowMaker.
As depicted in Fig. \ref{fig:short_case}, 
ShowMaker had significant improvements in generation and prediction performance,
by incorporating our ShowMaker designs.
Additionally, 
it can leverage visual prompts in the spatial actor cross attention,
for distinguishing the ``Teddy'' concept within text prompts.

\section{Conclusion}
\label{sec:conclusion}

In this paper, 
we introduce a generic system Vlogger, to generate over 5-minute vlogs from open-world descriptions, 
without loss of video coherence on script and actor.
Moreover, 
we present a novel video diffusion model ShowMaker for boosting state-of-the-art T2V generation and prediction. 
Finally,
we will release all the models, data and codes afterward,
allowing to develop further designs toward long video generation in the open world.
\appendix

\begin{table*}[h!]\centering
    \begin{minipage}{0.98\linewidth}\vspace{0mm}    
    \centering
    \begin{tcolorbox} 
        \centering
        \begin{tabular}{p{0.99\linewidth}}
        \begin{minipage}{0.99\linewidth}
User: The following is a story enclosed in three single quotes ```\textcolor[RGB]{243, 156, 18}{Once upon a time, there was a teddy bear named Teddy who dreamed of traveling the world. One day, his dream came true to travel around the world. Teddy sat in the airport lobby and traveled to many places of interest. Along the way, Teddy visited the Eiffel Tower, the Great Wall, and the pyramids. In Paris, Teddy had a picnic and enjoyed some delicious croissants. At the Great Wall of China, he climbed to the top and marveled at the breathtaking view. And in Egypt, he explored the pyramids and even found a secret treasure hidden inside. After his exciting journey, Teddy was eventually reunited with his owner who was thrilled to hear about all of Teddy's adventures. From that day on, Teddy always dreamed of traveling the world again and experiencing new and exciting things.}''' I would like to request your assistance in writing a script for a vlog based on this story. Provide the script in JSON format(do not answer anything else) with the following keys: video fragment id, video fragment description. You must answer like the content in the following three single quotes: ```[\{``video fragment id": 1, ``video fragment description": ``(the description, describe the characters, actions, and backgrounds in the video fragment)",\}, \{
``video fragment id": 2, ``video fragment description": ``(the description, describe the characters, actions, and backgrounds in the video fragment)",\}]'''  The descriptions of the video segments should adhere to the following guidelines: 1. Fits the original storyline. 2. All video fragment descriptions cannot conflict with each other, and the descriptions corresponding to successive fragments in the original story must have a certain continuity. 3. The description only needs to describe the visual elements presented, such as the subject, action, background, etc., and do not appear useless descriptions, such as mental activities.

GPT-4: \textcolor[RGB]{37, 204, 247}{[\{``video fragment id": 1, ``video fragment description": ``The screen opens to the sight of a well-loved teddy bear named Teddy. In the background, the hustle and bustle of an airport can be seen with people hurrying about to catch their flights."\}, \{``video fragment id": 2, ``video fragment description": ``The next scene shows Teddy looking out from the backpack, wide-eyed at the marvels of the Eiffel Tower in Paris. The iconic tower dominates the skyline, and Teddy is seen seated on a picnic blanket, with crumbs from a croissant nearby."\}, \{``video fragment id": 3, ``video fragment description": ``The setting shifts to the imposing stone structure of the Great Wall of China. Teddy is seen at the top of the wall, looking out over the breathtaking view. Views of the surrounding landscape are beautifully captured in wide shots."\}, \{``video fragment id": 4, ``video fragment description": ``In the next transition, we see Teddy exploring the mysterious pyramids of Egypt. He is captured in various poses around the ancient structures. A climactic moment occurs when Teddy discovers a secret treasure hidden deep within one of the pyramids."\}, \{``video fragment id": 5, ``video fragment description": ``The final scenes show Teddy being reunited with his owner. The joyous reunion takes place in a cozy home setting. The owner's surprised reaction upon hearing Teddy's adventures is clearly depicted. As the camera pulls away, Teddy sits happily in a familiar comforting spot, dreaming of more adventures."\}]
}
 \end{minipage}
        \end{tabular}
    \end{tcolorbox}
    \vspace{-2mm}
    \caption{\textbf{The Rough Script planning process of Vlogger.} The \textcolor[RGB]{37, 204, 247}{Rough Script} and \textcolor[RGB]{243, 156, 18}{Story} are marked by \textcolor[RGB]{37, 204, 247}{spiro disco ball} and \textcolor[RGB]{243, 156, 18}{orrange} respectively.
    }
    \label{tab:instruction_1}
\end{minipage}
\end{table*}
\begin{table*}[h!]\centering
    \begin{minipage}{0.98\linewidth}\vspace{0mm}    
    \centering
    \begin{tcolorbox} 
        \centering
        \begin{tabular}{p{0.99\linewidth}}
        \begin{minipage}{0.99\linewidth}
User: I want to make a vlog, the following is the original story enclosed in three single quotes ```\{\textcolor[RGB]{243, 156, 18}{Story}\}''' and this is my vlog production script in the following three single quotes ```\{\textcolor{cyan}{Rough Script}\}''', a paragraph in the script corresponds to a clip of the vlog, but the description of some video clips is too complicated, please help me analyze and rewrite a video script, split each description into several short descriptions and as more as possible. For example, if there is one paragraph in the script I gave you, then you should split it into fifteen paragraphs. Provide me the answer in JSON format(do not answer anything else) with the following keys: video fragment id, video fragment description. You must answer like the content in the following three single quotes: ```[\{``video fragment id": 1, ``video fragment description": ``(the description)",\}, \{``video fragment id": 2, ``video fragment description": ``(the description)",\}, \{``video fragment id": 3, ``video fragment description": ``(the description)",\}]'''. Remember to make sure that the description of each video clip is not long, no more than ten words, but there can be so many video clips. As many fragments as possible, as detailed as possible! Ensure that all description statements are as natural and syntactically correct as possible. Very important!!!: Try to have only one character in the description and avoid complex actions in the video fragment description, such as loaded in, fight, etc.

GPT-4: \textcolor{cyan}{[\{``video fragment id": 1, ``video fragment description": ``One day, Teddy was planning his trip on paper."\}, \{``video fragment id": 2, ``video fragment description": ``Backpack is in an airport."\}, \{``video fragment id": 3, ``video fragment description": ``People rush around the airport."\}, \{``video fragment id": 4, ``video fragment description": ``Teddy peeks out from backpack."\}, \{``video fragment id": 5, ``video fragment description": ``Eiffel Tower comes into view."\}, \{``video fragment id": 6, ``video fragment description": ``Teddy sits on a blanket."\}, \{``video fragment id": 7, ``video fragment description": ``Croissant crumbs lie nearby."\}, \{``video fragment id": 8, ``video fragment description": ``Scene shifts to Great Wall."\}, \{``video fragment id": 9, ``video fragment description": ``Teddy is at the wall's top."\}, \{``video fragment id": 10, ``video fragment description": ``Wide shot of surrounding landscape."\}, \{``video fragment id": 11, ``video fragment description": ``Transition to Egyptian pyramids."\}, \{``video fragment id": 12, ``video fragment description": ``Teddy poses near the structures."\}, \{``video fragment id": 13, ``video fragment description": ``A secret treasure is discovered."\}, \{``video fragment id": 14, ``video fragment description": ``Transition to cozy home setting."\}, \{``video fragment id": 15, ``video fragment description": ``Teddy reunites with his owner."\}, \{``video fragment id": 16, ``video fragment description": ``Owner reacts with surprise."\}, \{``video fragment id": 17, ``video fragment description": ``Teddy sits in a comfortable spot."\}, \{``video fragment id": 18, ``video fragment description": ``Teddy dreams of more adventures."\}
]
}
 \end{minipage}
        \end{tabular}
    \end{tcolorbox}
    \vspace{-2mm}
    \caption{\textbf{The Detail Script planning process of Vlogger.} The \textcolor{cyan}{Detail Script}, \textcolor[RGB]{37, 204, 247}{Rough Script} and \textcolor[RGB]{243, 156, 18}{Story} are marked by \textcolor{cyan}{cyan}, \textcolor[RGB]{37, 204, 247}{spiro disco ball}, and \textcolor[RGB]{243, 156, 18}{orrange} respectively.
    }
    \label{tab:instruction_2}
\end{minipage}
\end{table*}
\begin{table*}[h!]\centering
    \begin{minipage}{0.98\linewidth}\vspace{0mm}    
    \centering
    \begin{tcolorbox} 
        \centering
        \begin{tabular}{p{0.99\linewidth}}
        \begin{minipage}{0.99\linewidth}
User: The following is a story enclosed in three single quotes ```\{\textcolor[RGB]{243, 156, 18}{story}\}'''. I want to make a video according to this story, this is my video production script in the following three single quotes ```\{\textcolor{cyan}{Detail Script,}\}''', a paragraph in the script corresponds to a clip of the video. However, there may be some plots in the story missing, such as important plot missing, or transitions between pictures, please check and complete it for me. Provide me the answer in JSON format(do not answer anything else) with the following keys: video fragment id, video fragment description. You must answer like the content in following three single quotes: ```[\{``video fragment id": 1, ``video fragment description": ``(the description)",\}, \{ ``video fragment id": 2, ``video fragment description": ``(the description)",\}, \{``video fragment id": 3, ``video fragment description": ``(the description)",\}]'''. Remember to make sure that the description of each video clip is not long, no more than fifteen words, but there can be so many video clips. Each description should include the subject, place, and action as much as possible. As more fragment as possible, as detail as possible! Read this script carefully and don't pull down any details.

GPT-4: \textcolor[RGB]{52, 152, 219}{[\{``video fragment id": 1, ``video fragment description": ``Teddy is planning on paper."\}, \{``video fragment id": 2, ``video fragment description": ``The globe rotates with various locations in the background."\}, \{``video fragment id": 3, ``video fragment description": ``Teddy sitting in the airport lobby."\}, \{``video fragment id": 4, ``video fragment description": ``The plane took off from the runway."\}, \{``video fragment id": 5, ``video fragment description": ``Teddy arrives in Paris."\}, \{``video fragment id": 6, ``video fragment description": ``Teddy under the Eiffel Tower."\}, \{``video fragment id": 7, ``video fragment description": ``Teddy is enjoying a picnic."\}, \{``video fragment id": 8, ``video fragment description": ``Arrive at the Great Wall."\}, \{``video fragment id": 9, ``video fragment description": ``Teddy climbs the Great Wall."\}, \{``video fragment id": 10, ``video fragment description": ``Teddy watches the scenery from the Great Wall."\}, \{``video fragment id": 11, ``video fragment description": ``Teddy leaves the Great Wall."\}, \{``video fragment id": 12, ``video fragment description": ``Teddy is sitting on the bus, the scenery outside the window is beautiful."\}, \{``video fragment id": 13, ``video fragment description": ``Planes fly over the sky."\}, \{``video fragment id": 14, ``video fragment description": ``Teddy lands in a sandy landscape."\}, \{``video fragment id": 15, ``video fragment description": ``Pyramids gradually appeared in front of Teddy."\}, \{``video fragment id": 16, ``video fragment description": ``Teddy is entering the pyramid."\}, \{``video fragment id": 17, ``video fragment description": ``Teddy walks in the dark corridor."\}, \{``video fragment id": 18, ``video fragment description": ``Teddy suddenly walked into a bright room."\}, \{``video fragment id": 19, ``video fragment description": ``A treasure is revealed."\}, \{``video fragment id": 20, ``video fragment description": ``Teddy found his master with the treasure."\}, \{``video fragment id": 21, ``video fragment description": ``Room is filled with adventure mementos."\}, \{``video fragment id": 22, ``video fragment description": ``Teddy is looking at his mementos and camera rotation."\}, \{``video fragment id": 23, ``video fragment description": ``The globe gradually stopped rotating and the background gradually turned black."\}, \{``video fragment id": 24, ``video fragment description": ``Teddy sits in a chair and dreams about his next travels."\}]}

 \end{minipage}
        \end{tabular}
    \end{tcolorbox}
    \vspace{-2mm}
    \caption{\textbf{The Complete Script planning process of Vlogger.} The \textcolor[RGB]{52, 152, 219}{Complete Script}, \textcolor{cyan}{Detail Script} and \textcolor[RGB]{243, 156, 18}{Story} are marked by \textcolor[RGB]{52, 152, 219}{sea blue}, \textcolor{cyan}{cyan} and \textcolor[RGB]{243, 156, 18}{orrange} respectively.
    }
    \label{tab:instruction_3}
\end{minipage}
\end{table*}
\begin{table*}[h!]\centering
    \begin{minipage}{0.98\linewidth}\vspace{0mm}    
    \centering
    \begin{tcolorbox} 
        \centering
        \begin{tabular}{p{0.99\linewidth}}
        \begin{minipage}{0.99\linewidth}
User: The following is a story enclosed in three single quotes ```\{\textcolor[RGB]{243, 156, 18}{story}\}''', I want to make a video according to this story, this is my video production script in the following three single quotes ```\{\textcolor[RGB]{52, 152, 219}{Complete Script}\}''', a paragraph in the script corresponds to a clip of the video, Now that you know that 16-frame videos have a length of 2 seconds, please help me plan how much time it will take for each video clip to fully interpret the meaning of the script. Provide me the answer in JSON format(do not answer anything else) with the following keys: video fragment id, time. You must answer like the content in the following three single quotes: [\{``video fragment id": 1, ``time": 2,\}, \{``video fragment id": 2, ``time": 10,\}, \{``video fragment id": 3, ``time": 6,\}]

GPT-4: \textcolor[RGB]{30, 144, 255}{[\{``video fragment id": 1, ``time": 6\}, \{``video fragment id": 2, ``time": 10\}, \{``video fragment id": 3, ``time": 6\}, \{``video fragment id": 4, ``time": 7\}, \{``video fragment id": 5, ``time": 8\}, \{``video fragment id": 6, ``time": 6\}, \{``video fragment id": 7, ``time": 8\}, \{``video fragment id": 8, ``time": 9\}, \{``video fragment id": 9, ``time": 7\}, \{``video fragment id": 10, ``time": 8\}, \{``video fragment id": 11, ``time": 7\}, \{``video fragment id": 12, ``time": 10\}, \{``video fragment id": 13, ``time": 7\}, \{``video fragment id": 14, ``time": 6\}, \{``video fragment id": 15, ``time": 8\}, \{``video fragment id": 16, ``time": 5\}, \{``video fragment id": 17, ``time": 8\}, \{``video fragment id": 18, ``time": 6\}, \{``video fragment id": 19, ``time": 9\}, \{``video fragment id": 20, ``time": 6\}, \{``video fragment id": 21, ``time": 7\}, \{``video fragment id": 22, ``time": 8\}, \{``video fragment id": 23, ``time": 10\}, \{``video fragment id": 24, ``time": 6\}]}

 \end{minipage}
        \end{tabular}
    \end{tcolorbox}
    \vspace{-2mm}
    \caption{\textbf{The Script planning process of Vlogger.} The \textcolor[RGB]{30, 144, 255}{Script}, \textcolor[RGB]{52, 152, 219}{Complete Script} and \textcolor[RGB]{243, 156, 18}{Story} are marked by \textcolor[RGB]{30, 144, 255}{clear shill}, \textcolor[RGB]{52, 152, 219}{sea blue} and \textcolor[RGB]{243, 156, 18}{orrange} respectively.
    }
    \label{tab:instruction_4}
\end{minipage}
\end{table*}
\begin{table*}[t]
    \centering
    \setlength\tabcolsep{4.5pt}
    \resizebox{1\linewidth}{!}{%
    \begin{tabular}{l | c | c | c | c | c | c | c | c | c | c | c | c | c | c | c}
        \toprule
    \multicolumn{1}{c |}{ } & \multicolumn{3}{| c}{\textbf{Teddy Travel}} & \multicolumn{3}{| c}{\textbf{Alfie Work}} & \multicolumn{3}{| c}{\textbf{Tiger Moon}} & \multicolumn{3}{| c}{\textbf{Pete Skiing}} & \multicolumn{3}{| c}{\textbf{Anna Cooking}} \\
    \midrule
        \textbf{Method} & \textbf{CI ($\uparrow$)} & \textbf{CT ($\uparrow$)} & \textbf{VD ($\uparrow$)} & \textbf{CI ($\uparrow$)} & \textbf{CT ($\uparrow$)} & \textbf{VD ($\uparrow$)} & \textbf{CI ($\uparrow$)} & \textbf{CT ($\uparrow$)} & \textbf{VD ($\uparrow$)} & \textbf{CI ($\uparrow$)} & \textbf{CT ($\uparrow$)} & \textbf{VD ($\uparrow$)} & \textbf{CI ($\uparrow$)} & \textbf{CT ($\uparrow$)} & \textbf{VD ($\uparrow$)}  \\
        
        \midrule
        w/o \small{autoregressive}~\cite{Zhu2023MovieFactoryAM} & 0.5659 & 0.2846 & 48s & 0.5537 & 0.2641 & 1min20s & 0.6918 & 0.2819 & 1min10s & 0.4946 & 0.2627 & 54s & 0.5356 & 0.288 & 1min04s \\
        
        only \small{autoregressive}~\cite{villegas2023phenaki} & 0.5578 & 0.2554 & 2min47s & 0.5445 & 0.2361 & 5min12s & 0.6859 & 0.262 & 4min17s & 0.5033 & 0.2463 & 3min22s & 0.5295 & 0.2675 & 4min11s \\
        
        \textbf{Ours} & \textbf{0.653} & \textbf{0.2847} & \textbf{2min47s} & \textbf{0.6168} & \textbf{0.262} & \textbf{5min12s} & \textbf{0.7848} & \textbf{0.2802} & \textbf{4min17s} & \textbf{0.5393} & \textbf{0.2639} & \textbf{3min22s} & \textbf{0.553} & \textbf{0.2871} & \textbf{4min11s} \\
        \bottomrule
    \end{tabular}
    }
    \caption{\textbf{Detail ablation for generation process}.
    CI, CT, and VD refer to CLIP-I, CLIP-T, and Vlog Duration respectively.
    }
    \vspace{-1em}
    \label{tab:framework_ablation_detail}
\end{table*}
\begin{table*}[t]
\begin{minipage}[t]{1\linewidth}
    \centering
    \setlength\tabcolsep{6pt}
    \resizebox{1\linewidth}{!}{%
    \begin{tabular}{c | c | c | c | c | c | c | c | c | c | c | c | c | c | c | c | c | c | c | c | c | c }
    \toprule
    \multicolumn{2}{c |}{ } & \multicolumn{4}{| c}{\textbf{Experimental 1003}} & \multicolumn{4}{| c}{\textbf{Music 1179}} & \multicolumn{4}{| c}{\textbf{Animation 1175}} & \multicolumn{4}{| c}{\textbf{Sport 905}} & \multicolumn{4}{| c}{\textbf{Ads and Commercials 1229}} \\
    \midrule
        \textbf{TTCA} & \textbf{MTP} & 0/16 ($\downarrow$) & 1/15 ($\downarrow$) & 3/13 ($\downarrow$) & 5/11 ($\downarrow$) & 0/16 ($\downarrow$) & 1/15 ($\downarrow$) & 3/13 ($\downarrow$) & 5/11 ($\downarrow$) & 0/16 ($\downarrow$) & 1/15 ($\downarrow$) & 3/13 ($\downarrow$) & 5/11 ($\downarrow$) & 0/16 ($\downarrow$) & 1/15 ($\downarrow$) & 3/13 ($\downarrow$) & 5/11 ($\downarrow$) & 0/16 ($\downarrow$) & 1/15 ($\downarrow$) & 3/13 ($\downarrow$) & 5/11 ($\downarrow$)  \\
        \midrule
        \xmark & \xmark & 326.75 & 244.45 & 302.43 & 260.52 & 362.06 & 254.73 & 377.09 & 318.92 & 283.49 & 204.61 & 272.61 & 277.08 & 608.13 & 332.09 & 433.65 & 313.92 & 339.66 & 190.68 & 268.9 & 211.51 \\
        
        \cmark & \xmark & 319.93 & 197.03 & 251.3 & 226.34 & 357.28 & 206.5 & 261.78 & 221.05 & 252.42 & 174.8 & 262.87 & 243.34 & 556.09 & 261.2 & 315.4 & 218.27 & 332.7 & 177.28 & 205.07 & 166.19 \\
        
        \cmark & \cmark & \textbf{227.3} & \textbf{124.23} & \textbf{113.13} & \textbf{122.28} & \textbf{250.4} & \textbf{132.16} & \textbf{130.2} & \textbf{128.78} & \textbf{207.3} & \textbf{115.3} & \textbf{130.14} & \textbf{104.94} & \textbf{427.79} & \textbf{196.11} & \textbf{181.84} & \textbf{148.87} & \textbf{239.88} & \textbf{107.05} & \textbf{101.6} & \textbf{109.17} \\
        \bottomrule
    \end{tabular}
    }
    \subcaption{Ablation on Experimental, Music, Animation, Sport, and Ads and Commercials categories of Vimeo11K.}
    \end{minipage}

\begin{minipage}[t]{1\linewidth}
    \centering
    \setlength\tabcolsep{6pt}
    \resizebox{1\linewidth}{!}{%
    \begin{tabular}{c | c | c | c | c | c | c | c | c | c | c | c | c | c | c | c | c | c | c | c | c | c }
    \toprule
    \multicolumn{2}{c |}{ } & \multicolumn{4}{| c}{\textbf{Travel 1288}} & \multicolumn{4}{| c}{\textbf{Branded Content 1173}} & \multicolumn{4}{| c}{\textbf{Narrative 1010}} & \multicolumn{4}{| c}{\textbf{Comedy 1267}} & \multicolumn{4}{| c}{\textbf{Documentary 1064}} \\
    \midrule
        \textbf{TTCA} & \textbf{MTP} & 0/16 ($\downarrow$) & 1/15 ($\downarrow$) & 3/13 ($\downarrow$) & 5/11 ($\downarrow$) & 0/16 ($\downarrow$) & 1/15 ($\downarrow$) & 3/13 ($\downarrow$) & 5/11 ($\downarrow$) & 0/16 ($\downarrow$) & 1/15 ($\downarrow$) & 3/13 ($\downarrow$) & 5/11 ($\downarrow$) & 0/16 ($\downarrow$) & 1/15 ($\downarrow$) & 3/13 ($\downarrow$) & 5/11 ($\downarrow$) & 0/16 ($\downarrow$) & 1/15 ($\downarrow$) & 3/13 ($\downarrow$) & 5/11 ($\downarrow$)  \\
        \midrule
        \xmark & \xmark & 432.36 & 215.06 & 285.86 & 244.87 & 246.83 & 174.44 & 221.33 & 188.21 & 301.44 & 202.36 & 310.94 & 232.17 & 244.46 & 172.31 & 260.8 & 219.12 & 338.39 & 183.15 & 240.09 & 199.32 \\
        
        \cmark & \xmark & 382.39 & 178.86 & 220.89 & 172.71 & 249.88 & 135.09 & 173.19 & 150.78 & 272.07 & 173.55 & 224.69 & 192.73 & 232.65 & 135.15 & 191.19 & 182.11 & 380.86 & 143.28 & 199.26 & 161.65 \\
        
        \cmark & \cmark & \textbf{283.64} & \textbf{123.55} & \textbf{116.24} & \textbf{92.99} & \textbf{209.6} & \textbf{98.77} & \textbf{98.19} & \textbf{88.17} & \textbf{242.63} & \textbf{135.59} & \textbf{107.18} & \textbf{109.89} & \textbf{205.04} & \textbf{94.03} & \textbf{103.85} & \textbf{95.3} & \textbf{283.38} & \textbf{108.28} & \textbf{97.88} & \textbf{92.69} \\
        \bottomrule
    \end{tabular}
    }
    \subcaption{Ablation on Travel, Branded Content, Narrative, Comedy, and Documentary categories of Vimeo11K.}
    \end{minipage}
    
    \caption{\textbf{Detail ablation for temporal text cross attention and mixed training paradigm. }. Detailed experimental data on 10 major categories of Vimeo11K.
    }
    \vspace{-0.5cm}
    \label{tab:vimeo_detail}
\end{table*}

\section{Top-Down Planning}
\label{sub:planning}

As depicted in Tab. \ref{tab:instruction_1}, \ref{tab:instruction_2}, \ref{tab:instruction_3} and \ref{tab:instruction_4}, we encourage users to follow the provided example as guidance. 
These prompts instruct the LLM to generate structured outputs, specifically in the form of JSON code snippets. 
Concurrently, the examples furnished in these tables qualitatively emphasize the essentiality of a progressive approach in script generation, thereby rendering them qualitatively insightful.

\section{ShowMaker Implementation Details}
\textbf{Stage-1 Settings}.
In terms of architecture, we modify the input channel of the convolutional input layer in our base video pre-training model \cite{Wang2023LAVIEHV}, expanding it from 4 to 9. 
Furthermore, the integration of temporal text cross attention into the U-Net architecture results in a total parameter count of 1.3 billion. 
In terms of training, we train the entire U-Net, excluding only the spatial image cross attention components.
We set the value of the mixed training paradigm $\alpha$ and $m$ to 0.6 and 6 respectively. 
We establish the probability of dropping both the textual prompt and visual prompt at 0.1.
The learning rate we employ is 1e-4, and the batch size is 384, comprising a 16-frame video and 6 images per batch. 
We perform 54,000 steps total, requiring approximately five days to complete. 
It is worth noting that when training image data, the mask probability is set to 1, and the image features do not undergo temporal self attention and temporal text cross attention module.

\noindent\textbf{Stage-2 Settings}.
For architecture, we incorporate OpenCLIP ViT-H/14 \cite{ilharco2021openclip} as our image encoder. We set the spatial image cross attention $\beta$ to 1 during training.
For training, the U-Net is frozen while only permitting the training of spatial image cross attention components, thereby limiting the number of trainable parameters to only 22M. 
We follow a joint approach involving videos and images. 
However, unlike the first stage, we only need to train the video generation pattern in the second stage. 
During the training of spatial image cross attention, besides using a textual prompt as a condition, a visual prompt is also needed. 
The reference image for a given image is the image itself, while for videos, a random frame is selected from it. 
The probability of dropping both the textual prompt and visual prompt is set to 0.1. 
The learning rate used is 1e-4, and the batch size is 256, comprising a 16-frame video and 6 images per batch. 
We perform 62,000 steps, requiring approximately five days to complete. 

For both two stages, the AdamW \cite{Loshchilov2017DecoupledWD} optimizer is consistently employed. 
All images and videos used during training are subjected to online resizing, yielding a resolution of $320$$\times$$520$ resolution.
Regarding the video sampling process within the WebVid10M \cite{bain2021frozen}, we adopt a random sampling technique with an interval of 6 frames.

\section{Comparison with State-of-the-art}

\textbf{UCF-101} \cite{Soomro2012UCF101AD}.
To ensure accurate and stable computation of the FVD, we use the same sampling strategy as \cite{Wang2023LAVIEHV}, where a hundred videos are sampled for each class or hand-crafted prompt.
We first generate the video at a resolution of $320$$\times$$512$ and then resize it to match the $240$$\times$$320$ resolution of UCF-101.
Initially, videos are generated at a resolution of $320$$\times$$512$. 
Subsequently, these videos are resized to align with the $240$$\times$$320$ resolution of the UCF-101 dataset.
For the calculation of FVD \cite{unterthiner2019accurate}, we utilize a codebase derived from \cite{skorokhodov2022styleganv}.
\noindent\textbf{Kinetics-400} \cite{Kay2017TheKH}.
For Text-to-Video (T2V) generation within the Kinetics-400 dataset, we employ GPT-4 \cite{OpenAI2023GPT4TR} to construct a hand-crafted prompt that corresponds with each class label. 
Given the variable video resolutions present in the Kinetics-400 dataset, the ground truth videos are resized to a $320$$\times$$512$ resolution to match our generated videos.
\textbf{MSR-VTT} \cite{Chen2021TheMT}.
The MSR-VTT test set comprises 2,990 examples, each accompanied by 20 descriptions. 
For each example, we generate a video (comprising 16 frames at $320$$\times$$512$ resolution) using one randomly selected prompt. 
This procedure results in the creation of 2,990 videos.
The CLIPSIM \cite{Wu2021GODIVAGO} and FID \cite{parmar2022aliased} for the 47,840 frames are then computed following the methodology in \cite{singer2023makeavideo}.
\textbf{UCF-101 1000 Frames}.
In alignment with TATS \cite{Ge2022LongVG}, we use the hand-crafted prompts of UCF-101 as \cite{Ge2023PreserveYO} to generate videos and partition the generated long video into clips, each consisting of 16 frames. 
The FVD is computed between these clips and the ground truth.

\section{Ablation Study}
\label{sub:ablation}

\subsection{Vlog Generation Process}
As detailed in Tab. \ref{tab:framework_ablation_detail}, we used GPT-4 \cite{OpenAI2023GPT4TR} to automatically generate five scripts, namely \textit{Teddy Travel}, \textit{Alfie Work}, \textit{Tiger Moon}, \textit{Pete Skiing} and \textit{Anna Cooking} respectively.
Then we uniformly use the planning process of Vlogger to get the Script.
However, although we use the same Script, MovieFactory \cite{Zhu2023MovieFactoryAM} and Phenaki \cite{villegas2023phenaki} cannot use visual prompts due to their framework design, and can only use textual prompts for video generation.

\subsection{Temporal Text Cross Attention (TTCA) and Mixed Training Paradigm (MTP)}
As detailed in Tab. \ref{tab:vimeo_detail}, we get the video data of Vimeo11K from \href{https://vimeo.com/}{vimeo.com} and then use VideoChat \cite{Li2023VideoChatCV} to caption the videos.
Vimeo11K mainly contains 10 major categories of videos, namely \textit{Experimental}, \textit{Music}, \textit{Animation}, \textit{Sports}, \textit{Ads and Commercials}, \textit{Travel}, \textit{Branded Content}, \textit{Narrative}, \textit{Comedy}, and \textit{Documentary} respectively.
The number of videos included in each of the 10 categories is 1003, 1079, 1175, 905, 1229, 1288, 1173, 1010, 1267, and 1064 respectively, for a total of 11293.
The dataset and code for the test content will be released afterward.

\section{Visualization}
\label{sub:visualization}
We place the visualization results generated by our Vlogger in the supplementary material. 
Under the premise of not disrupting the files within the folder, please click on the following link: \href{https://zhuangshaobin.github.io/Vlogger.github.io/}{Project Page}.

\newpage
{
    \small
    \bibliographystyle{ieeenat_fullname}
    \bibliography{main}
}


\end{document}